\documentclass[10pt,twocolumn,letterpaper]{article}

\usepackage{cvpr}
\usepackage{times}
\usepackage{epsfig}
\usepackage{graphicx}
\usepackage{amsmath}
\usepackage{amssymb}

\usepackage[breaklinks=true,bookmarks=false]{hyperref}

\cvprfinalcopy % *** Uncomment this line for the final submission

\def\cvprPaperID{1} % *** Enter the CVPR Paper ID here

\begin{document}

%%%%%%%%% TITLE
\title{Dominant Codewords Selection with Topic Model for Action Recognition}

\author{Hirokatsu Kataoka, Kenji Iwata, Yutaka Satoh\\
National Institute of Advanced Industrial Science and Technology (AIST)\\
Tsukuba, Ibaraki, Japan\\
{\tt\small hirokatsu.kataoka@aist.go.jp}
\and
Masaki Hayashi, Yoshimitsu Aoki\\
Keio University\\
Hiyoshi, Yokohama, Kanagawa, Japan\\
{\tt\small }
\and
Slobodan Ilic\\
Technische Universitat Munchen (TUM)\\
Garching, Munich, Germany\\
{\tt\small }
}

\maketitle
%\thispagestyle{empty}

%%%%%%%%% ABSTRACT
\begin{abstract}
In this paper, we propose a framework for recognizing human activities that uses only in-topic dominant codewords and a mixture of intertopic vectors.
Latent Dirichlet allocation (LDA) is used to develop approximations of human motion primitives; these are mid-level representations, and they adaptively integrate dominant vectors when classifying human activities. In LDA topic modeling, action videos (documents) are represented by a bag-of-words (input from a dictionary), and these are based on improved dense trajectories (\cite{Wang2013b}). The output topics correspond to human motion primitives, such as finger moving or subtle leg motion. We eliminate the impurities, such as missed tracking or changing light conditions, in each motion primitive. The assembled vector of motion primitives is an improved representation of the action. We demonstrate our method on four different datasets.
\end{abstract}

\begin{figure*}[t]
\begin{center}
\includegraphics[width=170mm]{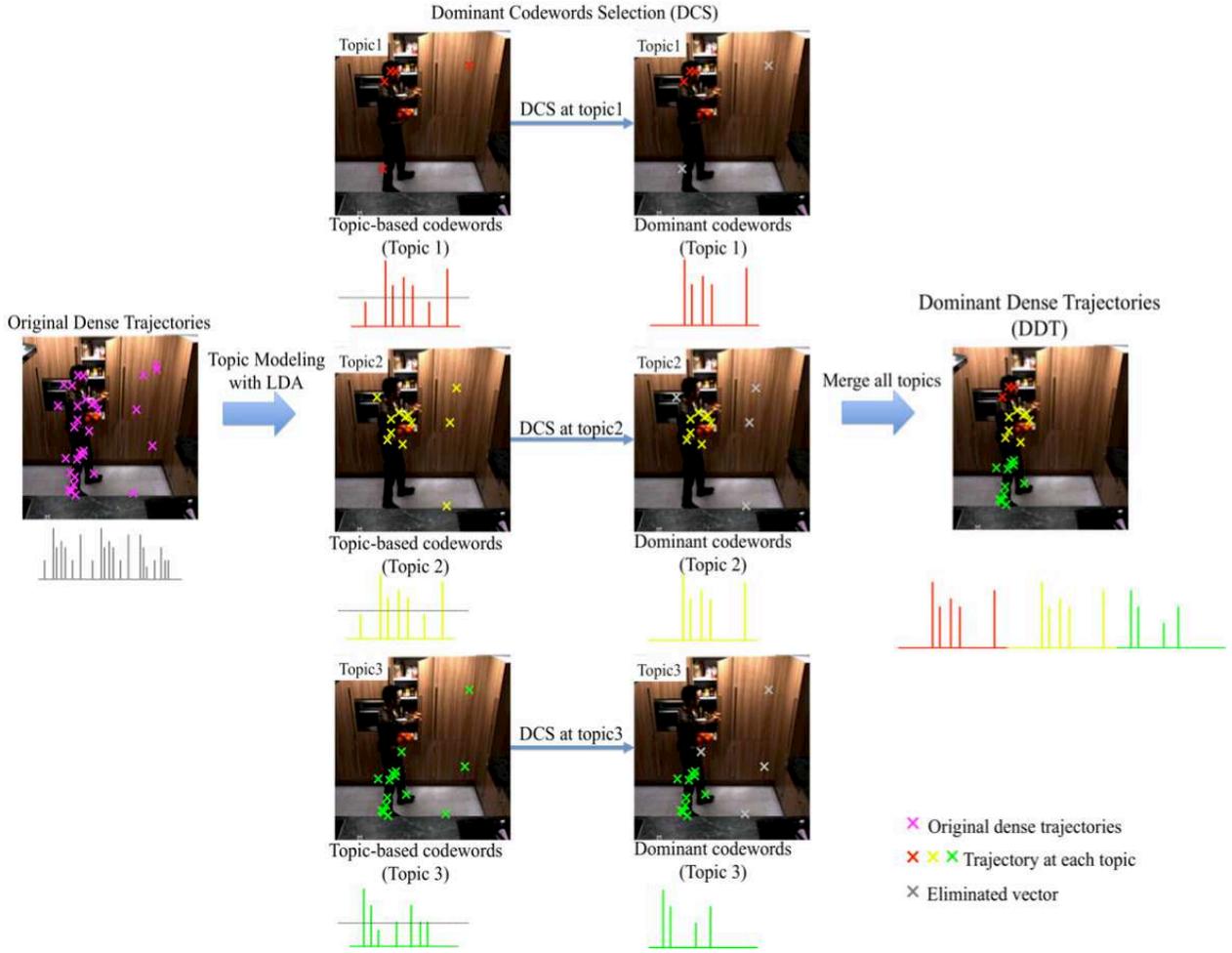}
\end{center}
\caption{The sampling points of dense trajectories (DT; left; Wang et al., 2013a) and our proposed framework, which uses dominant codewords selection (DCS), which is based on high-frequency codewords within a topic and the mixture vector of the dominant dense trajectories (DDT; right). In our proposal, the LDA analysis is used to recover the number of topics corresponding to the topic-based codewords (center left). The topic-based codewords represent the various motion primitives, such as movements of the head or legs, or swinging the arms. The dominant codewords are selected for the high-frequency elements of the various topic-based codewords, and these form the set of discriminative codewords (center right). Moreover, the DDT, which are a mixture vector of dominant codewords (right), allow us to accurately recognize human activities. The DCS reduces the noise in the DT samples of each topic.}
\label{figure:framework}
\end{figure*}

%%%%%%%%% BODY TEXT
\section{Introduction}
Various techniques for the visual analysis of human motion have been studied in the field of computer vision (Moeslund et al., 2011). Recently, human action recognition has become a very active research topic, and several survey papers have been published, including those of \cite{Aggarwal1999}, \cite{Moeslund2006}, and \cite{Aggarwal2011}. The number of applications is vast, and they include, but are not limited to, video surveillance, sports video analysis, medical imaging, robotics, video indexing, and games. To put these applications into practice, many action recognition methods have been recently proposed in order to improve accuracy. 

Current action recognition methods are able to process larger classes and more-complicated motions. The classic recognition pipeline (e.g., using space-time interest points; STIP) begins by extracting some kind of spatiotemporal features that are then fed to a classifier that has been trained to recognize such activities. However, in the case of human activities, minor differences between the extracted features frequently affect the classification of an action. This makes it difficult to use visual distinctions with existing feature descriptors. For human action recognition, \cite{Rohrbach2012} confirmed that dense trajectories (DT; Wang et al., 2013a) achieved better results than posture-based recognition. The most accurate state-of-the-art approach is a combination of improved DT (IDT)~\cite{Wang2013b} and per-frame deep-net features, such as the six-, seven-, and eight-layer approaches of~\cite{Krizhevsky2012}. According to the THUMOS challenge (Jiang et al., 2014), IDT should be used to accurately understand human activities. However, when doing so, the flow capture includes a large amount of noise, such as that due to incorrect tracking or changing light conditions. Such noise must be eliminated in order to obtain a dominant vector and attain more sophisticated action recognition.

In this paper, we propose the use of \textit{dominant codewords selection} (DCS) at each topic and the use of \textit{dominant dense trajectories} (DDT), which are an improved version of topic-based codewords for state-of-the-art fine-grained action recognition. In our proposed method, noise is eliminated from each motion primitive during the DCS, and the dominant codewords are combined to form the DDT. We use the bag-of-words (BoW) representation (Csurka et al., 2004) of the IDT (Wang and Schmid, 2013) to represent each action video with an action descriptor vector that contains the frequency of each of the various visual words. We then use latent Dirichlet allocation (LDA; Blei et al., 2003) to analyze the BoW and create topics. Intuitively, the topic-based codewords represent motion primitives from all the activities in the learning samples that are related to the movement of certain body parts, such as is shown in Figure~\ref{figure:framework} (center left). However, each topic still includes some noise due to incorrect tracking or changing light conditions. Therefore, we apply DCS in order to eliminate such noise from the IDT vectors. We are able to effectively eliminate the impurities for each topic (see Figure~\ref{figure:framework}; center right). Moreover, the DDT, which are a mixture of the dominant codewords using the AND operation, is effective for human action recognition (see Figure~\ref{figure:framework}; right). This vector is then used as input to a support vector machine (SVM). This framework is shown in Figure~\ref{figure:framework}. This model is a simple but effective way to create a dominant vector for action recognition.

We have added a contribution to this framework: we create a vector for human action recognition that is based on DCS and its mixture vector (DDT), and is based on topic modeling of a large number of visual words. The generic framework is a mixture vector of motion primitives, the DDT.

% Related works
\section{Related works}
The action recognition studies in the literature can be categorized into those that are historic and fine-grained and those that use topic-modeling approaches. 

\paragraph{Space-time action features} The first noteworthy work in action recognition proposed the use of STIP (Laptev, 2005). This idea extends the Harris corner detector into the time $t$ domain, and improvements using expanded feature representations were proposed in \cite{Laptev2008}, \cite{Marszalek2009}, and \cite{Everts2013}. As another way of representing spatiotemporal features, \cite{Yang2012} adopted a Gaussian mixture model to capture the frequency of each feature. In that approach method, primitive features are grouped separately when classifying activities. However, thus far, the best approach for action recognition is arguably the DT approach (Wang et al., 2013), which is based on descriptions of the trajectories of tracked feature points, which are densely sampled. When obtaining these trajectories, the following spatiotemporal features are used: the trajectory histograms of oriented gradients (HOG; Dalal and Triggs, 2005), histograms of optical flow (HOF; Laptev et al., 2008), and the motion boundary histograms (MBH; Dalal et al., 2006)].

Following the introduction of the original DT, dense sampling approaches for action recognition were also proposed in~\cite{Raptis2012, Jain2013, Peng2013, KataokaACCV2014, Wang2013b}. These studies improved the DT in various ways, for example, by introducing mid-level trajectory clustering (Raptis et al., 2013), eliminating extraneous flow (Jain et al., 2013), and extracting trajectories from within a motion boundary space (Peng et al., 2013). Kataoka \textit{et al.} (2014a) extended the co-occurrence feature descriptor and integrated it with the DT. They claimed that by adding dense sampling, they obtained more detailed descriptions, which resulted in better recognition performance. (Wang and Schmid (2013) obtained IDT by adding an estimate of the camera motion, detection-based noise canceling, and a Fisher vector (Perronnin et al., 2010).

Recent approaches have assigned human-object interactions to the IDT framework (Zhou et al., 2014, 2015). \cite{ZhouECCV2014} employed segmentation-based object information and weighted values with semantic-aware multiple kernel learning (SA-MKL), and \cite{ZhouCVPR2015} proposed binarized normed gradients (BING) objectness (Cheng et al., 2014) and human/object feature descriptions from object proposals. This mid-level approach with object information helps us to better understand human actions in context.

\paragraph{Topic modeling.}
\cite{Blei2003} proposed the use of LDA to improve the ad-hoc solutions provided by probabilistic latent semantic analysis (PLSA; Hoffmann, 1999). The LDA algorithm is based on a Bayesian framework, and it estimates latent topics by considering the distribution of documents. \cite{Niebles2008} extracted spatiotemporal features, represented them as a BoW, and then analyzed them with both PLSA (Hoffmann, 1999) and LDA (Blei et al., 2003). This approach determines the best topic to match each activity, and this allowed them to recognize several actions performed by multiple people in the input image, or to recognize single actions performed by a single person; in general, however, the actions must be different from each other. Compared to their approach, our proposal extracts representative activities as a mixture of motion primitives, and this allows a high degree of understanding of human actions. \cite{WangPAMI2009} used different BoW representations, and they assumed that latent topics directly corresponded to class labels. Their BoW representations are different from those in our approach, since in their approach, each frame in the action video is represented by a single visual word, whereas in our approach, each frame is represented by a collection of visual words. They claimed that human actions could be better characterized by large-scale features, rather than by local patches. 

However, effective subsequent use (Wang and Schmid, 2013) of DT of spatiotemporal features of patches has demonstrated their ability to produce high-precision action recognition.

\section{Framework for human action recognition}

In this section, we present our action recognition framework, which is based on topic modeling. This methodology is especially strong in classifying human activities.
Figure \ref{figure:framework} (right) shows the framework of motion primitives (dominant codewords) and its mixture vector (dominant BoW vector) for human action recognition.
The first step in action recognition is the extraction of BoW vectors based on the DT. 
These action features are then used as input to the LDA, which outputs the probability distribution of each topic, given the visual words. 
Topic modeling divides the input activities into the minimum units of action (motion primitive), e.g. finger moving or subtle leg motions. These motion primitives are then adaptively integrated into the dominant BoW vector, which is used as input to a SVM classifier.

% Vector calculation
\subsection{Feature extraction} 
The idea of DT (Wang et al., 2013) includes dense sampling of the image and extraction of spatiotemporal features from the trajectories of these features. In the feature extraction step, the DT adopts local feature descriptors for each image patch; these include the HOG, HOF, and MBH. Each patch comprises 32 pixels, which are divided into 2 $\times$ 2 blocks. The dimension of the HOG, HOF, and MBH are 96, 108, and 96, respectively. The final descriptions of the HOG and MBH are 2 (x) $\times$ 2 (y) $\times$ 3 (t) $\times$ 8 (directions). The HOF is 2 (x) $\times$ 2 (y) $\times$ 3 (t) $\times$ 9 (directions). In our implementation, the DT features are translated into 4,000 visual words for each feature descriptor.

\subsection{Dominant codewords selection with LDA}

LDA is a generative method for the analysis of documents and the distribution of the topics in a document. In our case, the documents are action videos, as represented by the action descriptors vector (the BoW representation of the DT features). Given these vectors as input, and with a defined number of topics $N_{T}$ as a parameter, the LDA algorithm outputs the probability distribution of the topics in the documents represented by the visual words (topic-based codewords).
More formally: let $V_i = \{v_{1}, v_{2}...v_{N}\}$ be an action descriptor vector corresponding to the action video $V_i, i=1,...,N$ represented by the set of visual words $vw_{j},j=1,...,N_{w}$ from the BoW vector, where $N_w$ is the number of codewords, i.e., the number of visual words in the dictionary (we note that in Wang et al., 2013a, $N_w$ = 4,000), and $N$ is the number of action videos in the training dataset. The learning vectors $V_{i}$ include all of the action classes that will be used to label the primitive motions in the activities. In order to apply LDA, $V_{i}$ should be cleaned by excluding nonzero values from the BoW, as follows: $\hat{V_{j}} = nonzero(vw_{j})$, where $\hat{V}={\hat{v_{1}}, \hat{v_{2}},...,\hat{v_{N_{w}}}}$ is the subset that includes only nonzero values; $\hat{N_{w}}$ ($\leq N_{w}$) is the size of the reduced action feature descriptor. These values are used as input to the LDA topic-modeling algorithm, and the output is the probability distribution of the topics given the visual words:
\begin{equation}
p(\Theta, \mathbf {T}| \mathbf{v}, \alpha, \beta) = \frac{p(\theta, \mathbf {T}, \mathbf{v} | \alpha, \beta)}{p( \mathbf{v} | \alpha, \beta)},
\end{equation}
where the topics are represented by $\mathbf{T}$ and $\Theta$ is the probability distribution of the topic. Here, $\alpha$ and $\beta$ are hyperparameters: $\alpha$ is the topic selection parameter, and $\beta$ selects the word when topic $T_{n}$ is selected.

We note that $p(\Theta, \mathbf {T} | \mathbf{v}, \alpha, \beta)$ can be simplified as follows: 
\begin{equation}
p(\theta,  \mathbf {T}_{n}|\gamma, \phi) = p(\Theta|\gamma)\prod_{n=1}^{N_{T}} p(T_{n}|\phi_{n}).
\end{equation}
We can estimate this distribution by using the Gibbs sampler (Griffiths and Steyvers, 2004), which is a Markov-chain Monte Carlo (MCMC) sampling algorithm. Note that the number of topics $N_{T}$ is a parameter of LDA. Moreover at each topic $T_{n}$ (topic-based codewords), the high-frequency codewords are selected to create a high-frequency motion primitive as $T'_{n}=Th(T_{n})$ (dominant codewords) with a frequency thresholding function $Th(*)$. The thresholding value is set to 1\% of the frequency of $N$.
Now, based on $p(\Theta, \mathbf {T} | \mathbf{v}, \alpha, \beta)$ and the action descriptor vector $V_i$, the action video can be represented as a collection of visual words belonging to the motion primitives (topics) contained in action video $i$. For example, if action video $i$ is represented by topics $T_{1}, T_{2}, T_{3}$ and each topic contains visual words $T_{1}=\{vw_{1},vw_{3},vw_{5}\}$, $T_{2}=\{vw_{1},vw_{2},vw_{7}\}$, and $T_{3}=\{vw_3,vw_6,vw_7\}$, by selecting the high-frequency motion primitives, we obtain $T'_1=\{vw_1,vw_5\}$, $T'_{2}=\{vw_{1},vw_{2},vw_{7}\}$, and $T'_{3}=\{vw_{7}\}$ (note that here, $vw_{3}$ and $vw_{6}$ are eliminated). By finding the union of the dominant codewords in all topics, we can represent the action video with the following topic feature descriptor: $VT_{i}=\{vw_1,vw_2,vw_5,vw_7\}$, which is the dominant BoW vector. Here, the $VT_{i}$ is represented as $VT_{i}=T_{1} \cap T_{2} \cap ... \cap T_{N_{t}}$. The dominant BoW vector $VT_{i}$ is the input to the SVM classifier.

\subsection{Extra feature descriptor for DCS and DDT}

Our general framework allows us to add extra features to the DCS. Here, we employ the co-occurrence feature suggested by \cite{KataokaACCV2014}. This approach adds an extended co-occurrence HOG (ECoHOG) (\cite{KataokaIEICE2014}) to the HOG, HOF, and MBH. The ECoHOG calculates the magnitude of edge pairs, as follows:
\begin{eqnarray}
   C_{x,y}(i,j) &=& \sum_{p=1}^{n}\sum_{q=1}^{m} \left\{ \begin{array}{ll} 
         \| g_{1}(p,q) \|+ \| g_{2}(p+x,q+y)\| & \;\; \\ \mathrm{if} \: d(p,q) = i \: \\ \mathrm{and} \: d(p+x, q+y) = j \\
        \\
        0 \;\;  \mathrm{otherwise}, \\
        \end{array}
        \right.
\end{eqnarray}
where $\|g(p,q)\|$ is the magnitude of the gradient, $C(i,j)$ is the co-occurrence histogram, and $i$ and $j$ quantize the orientation. The parameter tuning is based on that used by Kataoka et al. (2014a).

% Experiment
\section{Experiments}

\subsection{Four datasets for fine-grained activities}
\textbf{INRIA surgery dataset (INRIA; Huang et al., 2014).} This is a multiview dataset that including eight views and four fine-grained activities. Activities are performed by eight different people under various occlusions; e.g., people are occluded by a table or chair. The activities include \textit{cutting}, \textit{hammering}, \textit{repositioning}, and \textit{sitting}. Each person performed the same action twice, once for training and a second time for testing. Even though eight views are available, we used only four, which provided sufficient variety.

\begin{figure*}[t]
\begin{center}
\includegraphics[width=110mm]{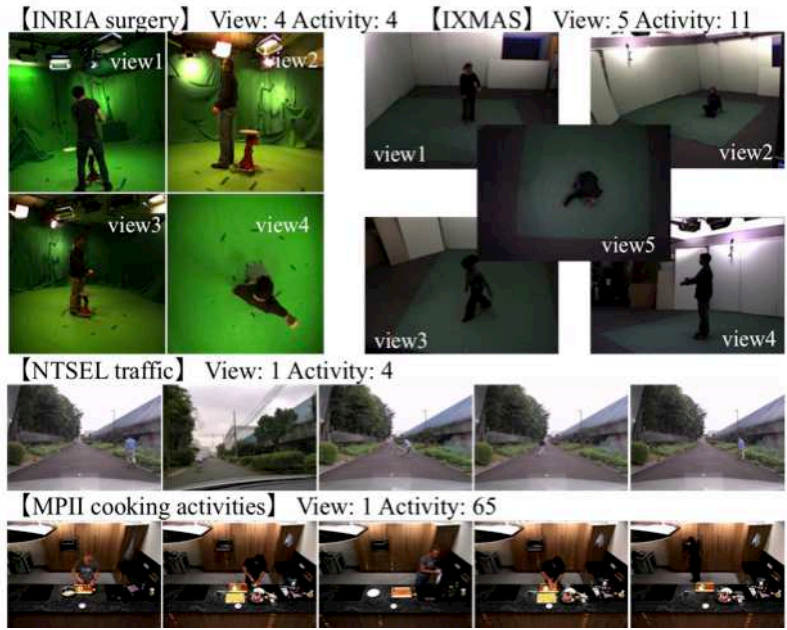}
\end{center}
\caption{The examples of the datasets that were used to evaluate our proposed method of action recognition: the INRIA dataset (Huang et al., 2014), the IXMAS dataset (Weinland eta al., 2007), the NTSEL dataset (Kataoka et al., 2015), and the MPII dataset (Rohrbach et al., 2012).}
\label{figure:dataset}
\end{figure*}

\textbf{IXMAS dataset (IXMAS; Weinland et al., 2007).} This multiview dataset includes five views and eleven different activities. Eleven activities are performed by ten people, each doing three trials. Cross-validation is executed to calculate the classification accuracy. The activities shown \textit{check watch}, \textit{cross arms}, \textit{scratch head}, \textit{sit down}, \textit{get up}, \textit{turn around}, \textit{walk}, \textit{wave}, \textit{punch}, \textit{kick}, and \textit{pick up}. 

\textbf{NTSEL traffic dataset (NTSEL; Kataoka et al., 2015).} \cite{KataokaITSC2015} collected $100$ videos of four activities that are forms of pedestrian motion in traffic scenes. The activities include \textit{walking}, \textit{crossing}, \textit{turning}, and \textit{riding a bicycle}. These videos show a cluttered background in small areas, and therefore, it is difficult to capture flows. The activities are fine grained, since there is little difference between \textit{walking}, \textit{crossing}, and \textit{turning}, all of which happen when the pedestrian is nearly stationary. We evaluated this dataset using five-fold cross-validation.

\textbf{MPII cooking activities dataset (MPII; Rohrbach et al., 2012).} This dataset is single view, and it contains 65 fine-grained activities performed by 12 participants. These activities can be broadly categorized into a few basic activities, such as seven ways of ``cutting" and five ways of ``taking". In all, the dataset comprises eight hours (881,755 frames) in 44 videos. Performance was evaluated by leave-one-person-out cross-validation.

% Setups
\subsection{Parameter setting}
\textbf{DT \& Classification.} The DT yield 4,000 dimensions represented by a BoW from each HOG, HOF, and MBH descriptor (we only used appearance features). The length of each trajectory was 15 frames. The classifier setting was based on the original DT (Wang et al., 2013a). 

\textbf{LDA.} The hyperparameters ($\alpha$, $\beta$) and the number of topic ($N_{t}$) should be preset when estimating feature distributions. Here, $\alpha$ and $\beta$ were set at 1.0 and 0.01, respectively (following Griffiths and Steyvers, 2004). $N_{t}$ was set to equal the number of activities in the dataset. This choice was not varied, and it gave good results in all experiments. The 1,000 Gibbs sampler iterations were performed to determine the sampling distribution. The threshold for selecting the dominant codewords from amongst the topic-based codewords was set at 1\% of the frequency of the learning vectors (e.g., the threshold was 50 when 5,000 sample vectors were input).

\textbf{Multiview setting.} For the multiview datasets (INRIA and IXMAS), LDA was used for topic modeling for each view separately, and afterwards, the results were combined. The range of topic modeling should be per view, since the appearance model slightly changes the settings between views. Therefore, the DDT were calculated in each view, and then the vectors were concatenated to construct a vector for multiple cameras.

\begin{table*}[t]
\caption{Comparison of our approach (LDA) to the baseline approach of DT (Wang et al., 2013a), using the INRIA, IXMAS, NTSEL, and MPII datasets.}
\begin{center}
\begin{tabular}{lcccccc}
Dataset & Type & HOG (\%) & HOF (\%) & MBHx (\%) & MBHy (\%) & Integ. (\%) \\

\hline\hline
INRIA & DT & 73.5 $\pm$ 0.4 & 75.8 $\pm$ 1.1 & 76.2 $\pm$ 2.5 & 65.9 $\pm$ 0.7 & 75.5 $\pm$ 0.3 \\
(4 views) & LDA & 74.4 $\pm$ 0.0 & 76.1 $\pm$ 0.4 & 76.7 $\pm$ 1.8 & 71.2 $\pm$ 0.7 & \textbf{80.4 $\pm$ 1.7} \\

\hline
IXMAS & DT & 86.5 $\pm$ 0.8 & 93.3 $\pm$ 0.1 & 91.4 $\pm$ 0.7 & 92.2 $\pm$ 0.1 & 93.1 $\pm$ 0.1 \\
(5 views) & LDA & 94.6 $\pm$ 0.3 & 90.2 $\pm$ 0.2 & 93.4 $\pm$ 0.6 & 93.9 $\pm$ 0.2 & \textbf{94.6 $\pm$ 0.3} \\

\hline
NTSEL & DT & 81.0 $\pm$ 3.1 & 89.1 $\pm$ 7.8 & 80.5 $\pm$ 6.9 & 76.6 $\pm$ 12.1 & 87.2 $\pm$ 4.9 \\
(1 view) & LDA & 79.6 $\pm$ 5.6 & 89.9 $\pm$ 7.9 & 80.5 $\pm$ 6.8 & 76.5 $\pm$ 11.9 & \textbf{90.9 $\pm$ 7.9} \\

\hline
MPII & DT & 54.3 $\pm$ 12.7 & 51.2 $\pm$ 6.6 & 51.1 $\pm$ 12.3 & 52.0 $\pm$ 13.5 & 59.5 $\pm$ 8.1 \\
(1 view) & LDA & 47.1 $\pm$ 13.3 & 58.5 $\pm$ 8.6 & 52.1 $\pm$ 13.6 & 52.7 $\pm$ 13.9 & \textbf{61.8 $\pm$ 8.3} \\

\hline
\end{tabular}
\end{center}
\label{table:dt_lda}
\end{table*}

% Results
\subsection{Experimental results}
\textbf{Comparison to the baseline (DT \& LDA).} The experimental results of the INRIA, IXMAS, NTSEL, and MPII datasets are shown in Table \ref{table:dt_lda}. The baseline is the DT accuracy, which was 75.5\%, 93.1\%, 87.2\%, and 59.5\% on the four datasets\footnote {Although our implementation is somewhat different from that of \cite{Wang2013a} and \cite{Rohrbach2012}, the accuracy of our results is close to that of their results (93.6\% for IXMAS and 59.2\% for MPII).}. The number of topics was set to correspond to the number of actions: 4 for INRIA, 11 for IXMAS, 4 for NTSEL, and 65 for MPII.
The results from Tables 1 through 3 indicate that the dominant BoW vector (proposed) outperforms the baseline DT for classifying actions. In Table \ref{table:dt_lda}, the proposed method achieved better accuracy, with values of 80.4\%, 94.6\%, 90.9\%, and 61.8\%. These values are +4.9\%, +1.5\%, +3.7\%, and +2.3\% better than the baseline, which extracted a 16,000-dimensional vector consisting of the HOG, HOF, MBHx, and MBHy (each of which is a 4,000-dimensional vector).
In the multiview implementation, the dominant BoW vector obtained after performing LDA significantly reduced the dimensionality: 30.8--59.0\% (6,371; 8,157; 4,936; and 9,447 dimensions) for INRIA, and 35.2--64.0\% for IXMAS (8,312; 5,640; 10,255; 9,023; and 9,970 dimensions), compared to the full action feature descriptor of the DT, which had 16,000 dimensions. Furthermore, 65.4\% and 77.4\% of the elements were used for classification in the fine-grained recognition scheme with NTSEL and MPII, which were implemented in single-view systems.
The INRIA dataset includes four types of action and is multiview; however, the MPII dataset has 65 activities. Undoubtedly, more elements are required to express various activities. LDA topic modeling is a significant representation, and it can extract topic-based codewords from both single- and multiview data. The dominant BoW vector is the mixture vector of dominant codewords. Thus, LDA excludes useless and noisy visual words, and a sophisticated feature space can more easily be built for the classification of fine-grained activities.

\begin{table}[t]
\caption{Comparison to the related feature selection approaches (top) to the state-of-the-art approaches (bottom). }
\begin{center}
\begin{tabular}{lcccc}

Feature Selection & & & & \\
\hline
Approach & INRIA & IXMAS & NTSEL & MPII \\
\hline \hline
DT + PCA & 79.0 & 94.1 & 90.6 & 60.2 \\
DT + ASS & 77.0 & 87.3 & 89.3 & 59.1 \\
DDT (Proposed) & \textbf{80.4} & \textbf{94.6} & \textbf{90.9} & \textbf{61.8} \\
\hline
 & & & & \\
State-of-the-art & & & & \\
\hline
Approach & INRIA & IXMAS & NTSEL & MPII \\
\hline \hline
\cite{Wang2013a} & 75.5 & 93.6 & 87.2 & 59.2 \\
\cite{Yang2012} & -- & 63.0 & -- & -- \\
\cite{Weinland2007} & -- & 91.1 & -- & -- \\
\cite{KataokaACCV2014} & -- & -- & -- & \textbf{62.4} \\
DDT (Proposed) & \textbf{80.4} & \textbf{94.6} & \textbf{90.9} & 61.8 \\
\hline
\end{tabular}
\end{center}
\label{table:comparison}
\end{table}

\begin{table}[t]
\caption{Comparison of the ECoHOG (Kataoka et al., 2014a) used with the DCS and a combination of the ECoHOG and the proposed method, using the MPII dataset.}
\begin{center}
\begin{tabular}{lc}
Approach (dimensions) & \% \\
\hline \hline
DDT (HOG, HOF, MBH) (12,388) & 61.8 \\
\hline
ECoHOG (4,000) (Kataoka et al., 2014) & 62.4 \\
\hline
DDT (ECoHOG) (1,598) & 65.6 \\
\hline
DDT: proposal (13,986)  & 68.9 \\
(ECoHOG, HOG, HOF, MBH) & \\
\hline
\cite{ZhouECCV2014}  & 70.5 \\
\hline
\cite{ZhouCVPR2015}  & \textbf{72.4} \\
\hline
\end{tabular}
\end{center}
\label{table:ecohog}
\end{table}

\begin{figure*}[t]
\begin{center}
\includegraphics[width=150mm]{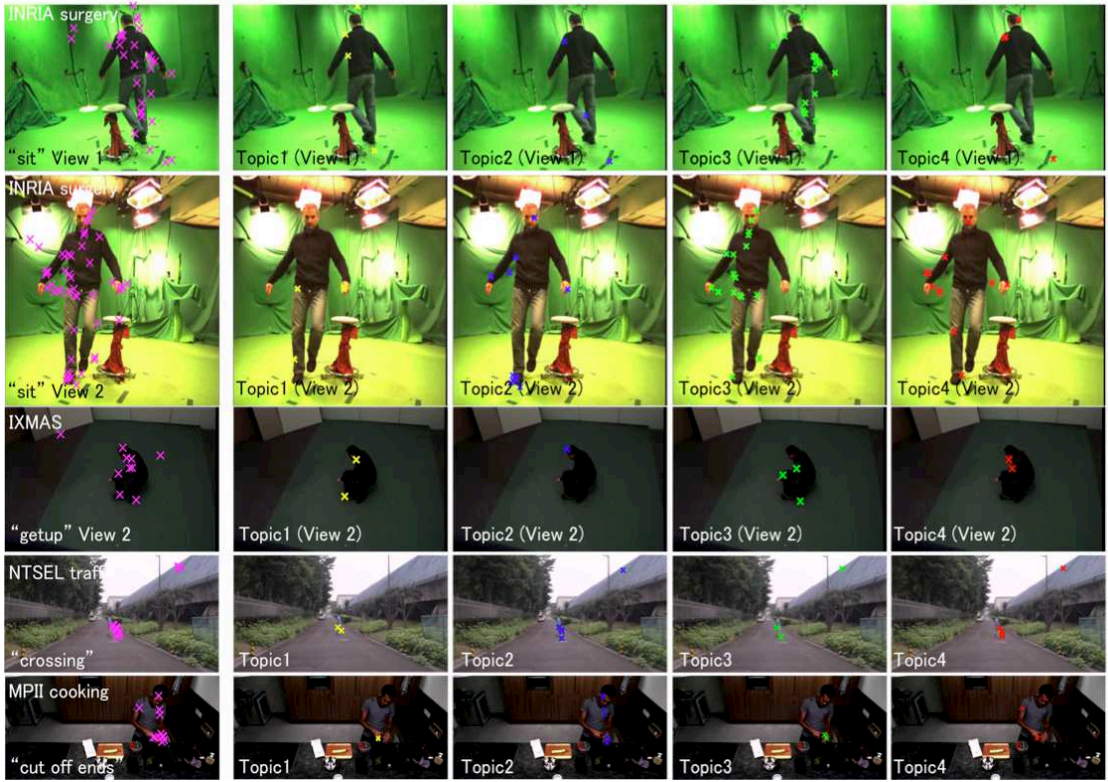}
\end{center}
\caption{Effect of feature pruning on topic modeling. Original DT features are shown in violet; features after topic modeling are shown in different colors for different topics. The topics are related to the high-frequency motion primitives.}
\label{figure:topicvisualization}
\end{figure*}

\textbf{Comparison to the subset vector approaches.} We considered the use of two approaches that reduce the dimensionality of the action feature vector in our proposed method; these were principal component analysis (PCA) and association mining (ASS; Quack et al., 2007). We obtained better results with these approaches, as shown in the upper part of Table~\ref{table:comparison}. This occurred even though ASS makes the smallest features with frequently used elements in the dataset. The use of frequent elements seems to enhance recognition, however, and only a few additional elements are needed to recognize sequential actions. We believe that our proposed method effectively estimates the distribution and creates sophisticated dominant BoW vectors from the action descriptor vectors. PCA results in a good representation, with results similar to that of our proposed method. The PCA representation frequently included useless visual words, such as those pertaining to the background or tracking noise. Parameter optimization, such as for the number of dimensions or for selecting the modeling samples, is time consuming. In our proposed method, the topic-based codewords are automatically generated by topic modeling, and therefore the selection of dominant codewords is simplified (it is only necessary to determine the elements with high frequency).

\textbf{Comparison to the state-of-the-art approaches.} Our LDA approach is compared to the state-of-the-art methods in the lower part of Table \ref{table:comparison}. The baseline is the DT result, as reported in \cite{Wang2013a}. We also compare our results to those of \cite{Weinland2007} and \cite{KataokaACCV2014} with the IXMAS and MPII datasets. Our method outperforms the state-of-the-art approaches. However, the ECoHOG of ~\cite{KataokaACCV2014} achieved the best result with the MPII dataset. 

\textbf{Co-occurrence feature combined with DCS.} We performed DCS with LDA on the ECoHOG features. The BoW vectorization was also performed. As expected, combining ECoHOG with LDA improved the results on the fine-grained MPII dataset. The results are shown in Table \ref{table:ecohog}. With the MPII dataset, the dominant BoW vector of ECoHOG regularly performed better than the raw vector, because of the sophisticated features modeled with LDA. ECoHOG has only 1,598 dimensions, compared to the 2,088 dimensions of the HOG, the 2,903 dimensions of the HOF, and the 3,760 and 3,637 dimensions in the $x$ and $y$ directions, respectively, of the MBH. The recognition rate was $65.6$\%, which was $3.2$\% better than when the raw ECoHOG (4,000 dimensions) was used. We integrated ECoHOG into the HOG, HOF, MBHx, and MBHy, and this combination of features achieved the highest performance, which was $68.9$\% accuracy with the MPII dataset. The recognition rate was the highest, and the dimensionality of the features was only $68.9$\% (13,986 dimensions = 1,598 $+$ 12,388) of the full 20,000 dimensions of the $5\times$ 4,000 BoW vectors. According to this result, the dominant BoW vector of ECoHOG and the original four DT features complement each other. Our proposal records competitive performance rate with \cite{ZhouECCV2014}, therefore, middle-level feature should be implemented in our framework.

\textbf{Topic visualization in single- and multiview data.} Figure \ref{figure:topicvisualization} shows the positions of the features extracted by DT (left column) and four topics decomposed with LDA (other columns). The violet crosses indicate the original DT features, and the other crosses correspond to the features belonging to the LDA topic feature descriptor associated with each motion primitive. Here, features belonging to different topics (high-frequency motion primitives) are shown in different colors: yellow for topic 1, blue for topic 2, green for topic 3, and red for topic 4. In these images, we can see that the noise due to incorrect flows in the background, shadows, and minor body movements has been removed. Moreover in the first two rows of Figure \ref{figure:topicvisualization}, the different features are captured from two different camera angles, and we can obtain an effective vector by using multiview concatenation. We obtained the following results for the separate views of the multiview datasets: INRIA: view1: 73.0\%, view2: 75.8\%, view3: 74.3\%, view4: 77.5\%; and IXMAS: view1: 92.9\%, view2: 93.2\%, view3: 94.4\%, view4: 93.4\%, view5: 94.0\%. These were combined to obtain the all-view concatenated vectors with the following results: INRIA: 80.4\%; and IXMAS: 94.6\%.

\section{Conclusion}
In this paper, we have proposed topic-based codewords and dominant vector selection for fine-grained action recognition. In evaluations on four different datasets (INRIA, IXMAS, NTSEL, and MPII), our method performed better than the original dense trajectories method and better than other related approaches. Moreover, we combined the extracted extended co-occurrence feature with the LDA and the original DT features. An outstanding performance, $68.9$\% in 65 cooking activities, was achieved by using five features (ECoHOG, HOG, HOF, MBHx, and MBHy) on the MPII dataset. We were also able to reduce the dimensionality of the feature vectors by approximately $70$\% (13,986 dimensions out of 20,000) on the MPII dataset. Intuitively, a topic describes a motion primitive from which noise (such as that due to background or shadows) has been removed. Minor body movements are also eliminated by the creation of the topic feature descriptors for each action. The use of high-frequency codewords and the associated mixture vector allows us to better understand fine-grained activities.

{\small
\bibliographystyle{ieee}
\bibliography{cvprw16}
}

\end{document}